\documentclass[letterpaper, 10 pt, conference]{ieeeconf}  

\IEEEoverridecommandlockouts                              

\usepackage{graphicx} 
\usepackage{amsmath}
\usepackage[table]{xcolor}
\usepackage{colortbl}
\usepackage{multirow}
\usepackage{multicol}
\usepackage{dblfloatfix} 
\usepackage{subcaption}
\usepackage{cite}
\usepackage{booktabs}
\usepackage{url}
\let\labelindent\relax
\usepackage{enumitem}

\usepackage{amssymb}  

\title{\LARGE \bf
SCDP: Learning Humanoid Locomotion from Partial Observations via Mixed-Observation Distillation
}

\author{
\authorblockN{Milo Carroll, Tianhu Peng, Lingfan Bao, Chengxu Zhou and Zhibin Li}
\authorblockA{Department of Computer Science, University College London\\
\texttt{\{milo.carroll.23, tianhu.peng.24, lingfan.bao.24, chengxu.zhou, alex.li\}@ucl.ac.uk}}
}

\makeatletter
\def\bstctlcite{\@ifnextchar[{\@bstctlcite}{\@bstctlcite[@auxout]}}
\def\@bstctlcite[#1]#2{\@bsphack
  \@for\@citeb:=#2\do{%
    \edef\@citeb{\expandafter\@firstofone\@citeb}%
    \if@filesw
      \immediate\write\csname #1\endcsname{\string\citation{\@citeb}}%
    \fi
  }%
  \@esphack}
\makeatother

\begin{document}

\bstctlcite{IEEEexample:BSTcontrol}

\maketitle
\thispagestyle{empty}
\pagestyle{empty}

\begin{abstract}

Distilling humanoid locomotion control from offline datasets into deployable policies remains a challenge, as existing methods rely on privileged full-body states that require complex and often unreliable state estimation. We present \textbf{Sensor-Conditioned Diffusion Policies }~(SCDP) that enables humanoid locomotion using only onboard sensors, eliminating the need for explicit state estimation. SCDP decouples sensing from supervision through mixed-observation training: diffusion model conditions on sensor histories while being supervised to predict privileged future state–action trajectories, enforcing the model to infer the motion dynamics under partial observability. We further develop restricted denoising, context distribution alignment, and context-aware attention masking to encourage implicit state estimation within the model and to prevent train–deploy mismatch. We validate SCDP on velocity-commanded locomotion and motion reference tracking tasks. In simulation, SCDP achieves near-perfect success on velocity control (99–100\%) and 93\% tracking success in AMASS test set, performing comparable to privileged baselines while using only onboard sensors. Finally, we deploy the trained policy on a real G1 humanoid at 50 Hz, demonstrating robust real robot locomotion without external sensing or state estimation.


\end{abstract}


\section{INTRODUCTION}
\label{sec:intro}

Humanoid robot control has advanced rapidly in recent years, progressing from task-specific reinforcement learning controllers~\cite{margolis2022walktheseways, learntowalk, Tianhu2} to motion reference tracking methods~\cite{deepmimic, onmih20, humanplus, Bao_2025}. These approaches enable diverse whole-body behaviors and have supported applications such as real-time teleoperation and multi-skill locomotion in both graphics~\cite{maskedmimic} and robotics~\cite{onmih20, humanplus, twist, unitracker, gmt}. The ability to reproduce large skillets within a single controller is essential for adaptive behaviors, as demonstrated by multi-expert learning that synthesizes adaptive locomotion by combining different motor skills to handle varying task conditions~\cite{yang2020multi}.

Recently, diffusion models have become promising for learning more expressive control by modeling full distributions of expert behavior rather than deterministic action mappings~\cite{mdm, closd, bao_2025_2}. Their flexibility offers task adaptation via direct conditioning or classifier guidance~\cite{classGuidance, guidedMDM}. In robotics, diffusion-based policies have demonstrated strong performance in visuomotor manipulation~\cite{diffpolicy}, bipedal locomotion~\cite{diffuceloco, birodiff}, and physics-based humanoid control~\cite{pdp}. Guided state-action diffusion approaches for humanoids~\cite{diffusecloc, beyondmimic} further show the ability to perform look-ahead planning and multitask control within a unified framework.

Despite these advances, current humanoid diffusion controllers rely fundamentally on privileged full-body state information at deployment. These methods assume access to global position, orientation, base velocity, and rigid-body position, requiring complex and often unreliable state estimation pipelines on real robots outside the lab setting. The removal of privileged inputs dramatically degrades performance, with catastrophic failures even in simple tasks such as joystick locomotion~\cite{beyondmimic}. This dependence is not viable in most practical deployment scenarios, as only direct onboard proprioceptive sensing is reliably available. Learning solely from onboard sensing induces a partially observable Markov decision process (POMDP), in which policies must infer hidden global states from incomplete observations. Existing diffusion-based controllers are not designed to operate under partial observability.

We address this limitation with our proposed framework, SCDP, a diffusion-based distillation framework for humanoid control using only onboard proprioceptive measurements. Our key idea is to decouple sensing from supervision via mixed-observation training. During training, the model conditions only on onboard sensor histories while being supervised to predict privileged full-body state trajectories. This asymmetry enforces internal representation learning that captures global body dynamics from partial sensing, eliminating explicit state estimation at deployment.

\begin{figure}[t]
    \centering
    \includegraphics[width=1.0\linewidth]{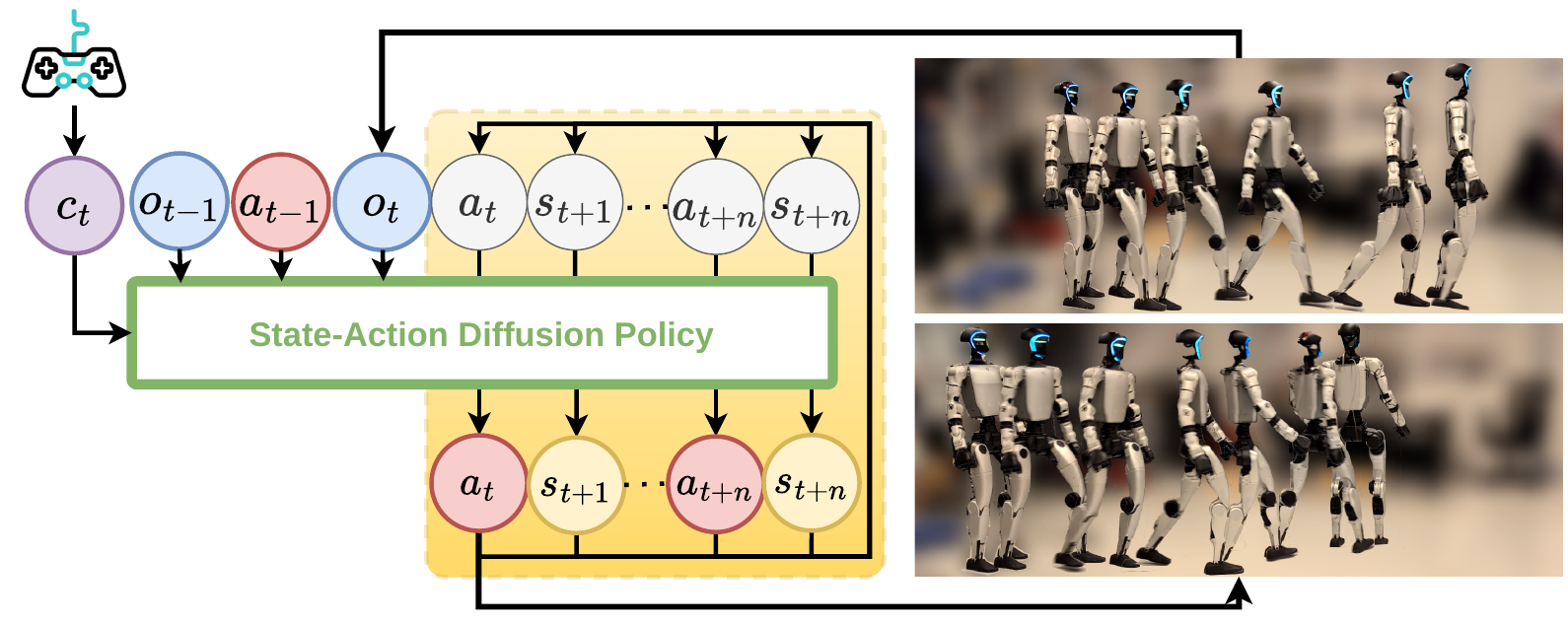}
    \caption{Deployment of Sensor-Conditioned Diffusion Policies (SCDP) on Unitree G1, performing robust locomotion at 50~Hz using only proprioceptive sensors, without external motion capture or state estimation.}
    \label{fig:deployments}
\end{figure}

Unlike prior teacher–student transfer methods developed for MLP-based locomotion policies~\cite{learntowalk, dagger, onmih20, xbody2}, which distill reactive mappings from observation to action, SCDP addresses a fundamentally different problem: distilling a generative trajectory planner from offline datasets. While recent work has explored distilling RL policies into interpretable models for locomotion~\cite{distilling_2024}, our goal here is to enable such distillation under partial observability without privileged state information, where the student policy must jointly denoise temporally \textit{coherent state-action sequences} with incomplete information. Our proposed training formulation resolves this and enables deployment without privileged sensing.

\begin{figure*}[!t]
    \centering
    \includegraphics[width=\linewidth]{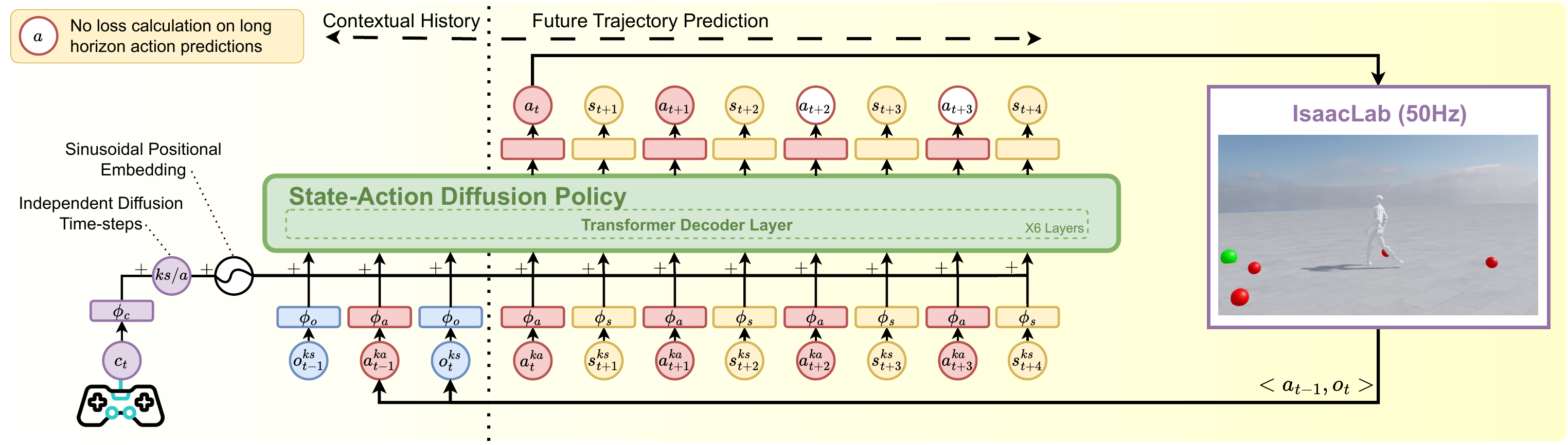}
    \caption{
        \textbf{Sensor-Conditioned Diffusion Policies (SCDP) architecture and training framework.} 
    The state-action diffusion policy conditions on sensor observation history $\{o_t\}$, past actions $\{a_t\}$, and commands $c_t$, while predicting future trajectories containing privileged states $\{s_t\}$ and actions. This mixed-observation formulation enables learning global dynamics from partial sensing. 
}
\label{fig:diffusion_arch}
\end{figure*}

The key contributions of this work are as follows:

\begin{itemize}[leftmargin=*]
    \item Mixed observation training, which conditions on sensor histories while predicting privileged state trajectories, enables implicit inference of global body dynamics.
    
    \item Restricted denoising, which excludes base velocity from denoising inputs while retaining it in supervision, forces the model to infer velocity from context, enabling control without velocity feedback.

    \item Context distribution alignment, which maintains consistent causal relationships between states and actions across training and inference.

    \item Context-aware attention masking enables bidirectional attention within the context window to facilitate inference of latent dynamics from partial observations.
   
    \item Experimental validation via deployment on physical Unitree G1 at 50~Hz (Fig.~\ref{fig:deployments}), and systematic ablations on data collection, context length, and architectural choices.
\end{itemize}

\section{Method}
\label{sec:method}

We present SCDP, a framework for distilling a diffusion-based controller from an expert RL policy to achieve whole-body humanoid control from onboard sensing. Our approach consists of four key components: (1) a Multi-Motion Tracking Policy (MMP) serving as an expert demonstrator; (2) our diffusion distillation framework; (3) a data collection methodology for robust distillation; and (4) a real-time deployment configuration for the Unitree G1. We describe each component below, showing how they combine to enable sensor-based control.



\subsection{Multi-Motion Tracking Policy}
\label{sec:mmp}
The foundation of our distillation framework is a Multi-Motion Policy (MMP), an RL-trained expert for multi-motion tracking that is built upon the reward function and PD controller from~\cite{beyondmimic}. All quantities are expressed in the anchor-centric (set as the torso), heading-aligned frame via $R_{\mathrm{yaw}}$, the rotation aligning the global frame with the root heading. For each tracked body $b \in \mathcal{B}$, we compute local positions, orientations, and velocities as:
\begin{equation}
\begin{split}
\tilde{p}_b = R_{\text{yaw}} (p_b - p_{\text{anchor}}), \quad
\tilde{R}_b = R_{\text{yaw}} R_b, \\
\tilde{v}_b = R_{\text{yaw}} v_b, \quad
\tilde{\omega}_b = R_{\text{yaw}} \omega_b
\end{split}
\end{equation}
$o_{\text{robot}} = [h_{\text{anchor}},\, \{\tilde{p}_b, \tilde{R}_b, \tilde{v}_b, \tilde{\omega}_b\}_{b \in \mathcal{B}},\, a_{t-1}]$ is the self-observation. The task observation provides both absolute and relative future states for each reference body $b_m \in \mathcal{B}_m$ over a look-ahead horizon of $H$ steps:
\begin{equation}
    o_{\text{task}} = \big[\{\hat{p}_{b_m}^{(k)},\, \Delta p_{b_m}^{(k)},\, \hat{R}_{b_m}^{(k)},\, \Delta R_{b_m}^{(k)}\}_{k=1}^{H}\big],
\end{equation}
where $\hat{p}_{b_m}^{(k)} = R_{\text{yaw}}(p_{b_m}^{(k)} - p_{\text{anchor}})$ and $\Delta p_{b_m}^{(k)} = R_{\text{yaw}}(p_{b_m}^{(k)} - p_b)$ capture absolute and relative positions respectively, with orientations defined analogously.

\subsection{Mixed-Observation Distillation: Learning Global Dynamics from Partial Sensing}

The key idea of mixed-observation distillation is to \emph{decouple sensing from supervision} during training. In particular, we condition the diffusion model on historical observations from onboard sensors ($o_t$), while supervising it to predict future trajectories that include privileged full-body states ($s_t$).
This asymmetry, shown in Fig.~\ref{fig:diffusion_arch}, enables the model to learn an implicit forward model of global body dynamics, allowing deployment without explicit state estimation.

\subsubsection{Problem Formulation}
We consider a humanoid robot with a full state $s_t \in \mathcal{S}$ and an action space $a_t \in \mathcal{A}$. The full state comprises:
$
s_t = \big[
        o_t,\;
        \{\tilde{p}_b\}_{b \in \mathcal{B}},\;
        \{\tilde{v}_b\}_{b \in \mathcal{B}}
    \big],
\label{eq:full_state}
$
where $o_t$ contains proprioceptive measurements and $\{\tilde{p}_b\}_{b \in \mathcal{B}}$, $\{\tilde{v}_b\}_{b \in \mathcal{B}}$ are the local positions and velocities of rigid bodies $b \in \mathcal{B}$ in the heading-aligned frame. These geometric quantities are available in simulation but not on hardware.

At inference, the robot accesses only onboard measurements:
$
  o_t = \big[
        v_{\text{pelvis}},\;
        \omega_{\text{pelvis}},\;
        q - q_{0},\;
        \dot{q},\;
        g_{\text{pelvis}},\;
        g_{\text{torso}}
    \big],
\label{eq:onboard_obs}
$
denoting pelvis linear velocity $v_{\text{pelvis}}$, pelvis angular velocity $\omega_{\text{pelvis}}$, joint positions relative to the default $q - q_0$, joint velocities $\dot{q}$, and gravity vectors in the pelvis and torso frames $g_{\text{pelvis}}, g_{\text{torso}}$. Note that $v_{\text{pelvis}}$ is excluded in our final model because velocity estimates are unreliable on hardware. Critically, $o_t$ lacks geometric information about body positions and provides only local kinematic and inertial measurements.

The diffusion policy $\pi(\tau_t \mid O_t, c_t)$ generates state-action trajectories conditioned on observation history $O_t$ and task commands $c_t$ (velocity or motion reference commands), where trajectories are defined as
$
\tau_t = [\, a_t,\; s_{t+1},\; a_{t+1},\; \ldots,\; s_{t+H} \,]
\label{eq:trajectory}
$
and observation histories are defined as 
$
O_t = [\, o_{t-N},\; a_{t-N},\; \ldots,\; o_{t-1},\; a_{t-1},\; o_t \,].
\label{eq:history}
$
Crucially, while \emph{inference} uses only $O_t$, \emph{training} supervises predictions of privileged states $s_{t+1:t+H}$. This mixed-observation formulation is the key to transferring full-body motion knowledge to sensor-only deployment.

\paragraph{Diffusion Model Formulation}
We model the problem using a denoising diffusion probabilistic model~(DDPM)~\cite{ddpm} to iteratively denoise trajectory samples. At diffusion timestep $k$, a noised trajectory $\tau_t^k$ is refined by predicting the clean trajectory $x_{0,\theta}(\tau_t^k, O_t, c_t)$:
\begin{equation}
\mathcal{L}_{\text{DDPM}} = \mathbb{E}_{\tau_t, k}\!\left[\, \|\, x_{0,\theta}(\tau_t^k, O_t, c_t) - \tau_t \,\|^2 \,\right].
\label{eq:ddpm_loss}
\end{equation}
Sampling proceeds by iteratively applying:
\begin{equation}
\tau_{t}^{k-1} = \alpha_k\!\left(\tau_{t}^{k} - \gamma_k\,\epsilon_\theta(\tau_{t}^{k}, O_t, c_t)\right) + \mathcal{N}(0,\sigma_k^2 I),
\label{eq:ddpm_sample}
\end{equation}
with standard DDPM coefficients $(\alpha_k, \gamma_k, \sigma_k)$ and predicted noise $\epsilon_\theta$ derived from $x_{0,\theta}$. We use independent noise schedules for state and action components of $\tau_t$ as in~\cite{diffusecloc}.

\subsubsection{Restricted Denoising}

A remaining challenge is achieving velocity control without direct velocity feedback. While mixed-observation training provides geometric information through prediction targets, the model can use partially-noised velocity terms as shortcuts during training. To force genuine velocity inference from context, we introduce restricted denoising which omits pelvis linear velocities $v_{\text{pelvis}}$ from the denoising inputs but retains them in the supervision target: $s_t^{\text{res}} = s_t \setminus \{v_{\text{pelvis}}\}$. The corresponding restricted trajectory is $\tau_t^{\text{res}} = [\, a_t,\; s_{t+1}^{\text{res}},\; a_{t+1},\; \ldots,\; s_{t+H}^{\text{res}} ]$.

The denoising model takes restricted trajectories as input but predicts full trajectories including velocities:
\begin{equation}
\mathcal{L}_{\text{DDPM}_{\text{res}}} = \mathbb{E}_{\tau_t, k}\!\left[\, \|\, x_{0,\theta}(\tau_t^{\text{res},k}, O_t, c_t) - \tau_t \,\|^2 \,\right].
\label{eq:restricted_input}
\end{equation}
This enforces the model to infer pelvis velocities from the historical context, developing robust velocity estimation without direct feedback.

\subsubsection{Context Distribution Alignment}
Prior works~\cite{diffusecloc, beyondmimic} train on $(s_{\text{noised}}, a)$ pairs, where $s_{\text{noised}}$ results from executing noised actions $a_{\text{noised}}$, creating a distribution mismatch at inference when the model observes clean, $(s, a)$ context. During training, we provide the context using $(s_{\text{noised}}, a_{\text{noised}})$ pairs. This aligns the training context and deployment conditions, reducing distribution shift and maintaining a causal relationship between $(s,a,s')$ of the context in both settings.

\subsubsection{Context-Aware Attention Masking}
Attention masking approaches in prior works~\cite{diffusecloc, beyondmimic} apply causal masks that prevent states from attending to actions, while allowing attention to future states. In contrast, we enable bidirectional attention within the context window while maintaining causal constraints only in the prediction horizon. This enables the model to aggregate historical information bidirectionally, facilitating the inference of latent dynamics from partial observations.

\subsubsection{Representation Alignment}
\label{sec:alignment}
We experiment with a cosine-similarity alignment loss to bridge the information gap between sensor observations and privileged states. Specifically, the loss encourages alignment between the sensor encoder output $\phi_o(o_t)$ and the privileged state encoder output $\phi_s(s_t)$, averaged over the context window. However, we found this approach detrimental to performance in our experiments (see Sec.~\ref{sec:feature_ablation}) and excluded it from the final SCDP model.

\subsubsection{Velocity Conditioning}
We condition on desired pelvis velocities $c_t = [v^x_{\text{cmd}}, v^y_{\text{cmd}}, \dot{\psi}_{\text{cmd}}] \in \mathbb{R}^3$ via MLP encoding $\phi_c: \mathbb{R}^3 \to \mathbb{R}^d$ combined with timestep embeddings: $e_t = e^{\text{time}}_t + \phi_c(c_t)$. To enforce tracking, we minimize $\mathcal{L}_{\text{cmd}} = \|c_t - \bar{c}_t\|_2^2$ where $\bar{c}_t$ is the mean predicted velocity over the horizon.

\subsubsection{Motion Reference Conditioning}
For motion tracking, we condition on $c_t = [q_{\text{ref}}, \dot{q}_{\text{ref}}, \mathbf{R}_{\text{rel}}]$, where $\mathbf{R}_{\text{rel}} = \mathbf{q}_{\text{robot}}^{-1} \otimes \mathbf{q}_{\text{ref}}$ represents the relative orientation between robot and reference frames. The complete training objective is:
\begin{equation}
\mathcal{L} = \mathcal{L}_{\text{DDPM-res}} + \lambda_{\text{cmd}}\,\mathcal{L}_{\text{cmd}}.
\label{eq:total_loss}
\end{equation}

\subsection{Datasets}
\label{sec:datasets}
The MMP is trained on walking motions from PHC-filtered AMASS (AMASS-walk)~\cite{amass, onmih20}, retargeted via Mink~\cite{mink}. For SCDP distillation, we select 65 diverse clips, mirror them (130 total), and use the expert to collect 40 rollouts per reference with stochastic actions~\cite{pdp}, domain randomization~\cite{beyondmimic}, and push forces (every 1-3s). This yields 5200 trajectories ($\sim$750k timesteps at 50~Hz). Motion tracking uses the full AMASS-walk training set, with only a single rollout per trajectory.

\subsection{Training and Deployment Configuration}
\label{sec:implementation}
The MMP is a six-layer MLP (hidden dims: [1024, 1024, 1024, 1024, 512, 128]) trained using PPO~\cite{ppo} for 100k iterations in IsaacLab~\cite{isaaclab} across 4096 parallel environments with a learning rate of $3 \times 10^{-4}$. Training takes approximately four days on an NVIDIA RTX 5090.

The SCDP diffusion model uses a six-layer Transformer encoder with four attention heads, embedding dimension 512, and feed-forward dimension 1024. Four encoders process actions (1-layer MLP), onboard observations (2-layer MLP), privileged states (2-layer MLP), and commands (2-layer MLP). The model predicts 16-step trajectories with action losses weighted to zero beyond 8 steps for near-term accuracy; at inference, only the first action is executed before replanning. Training uses 1M gradient updates with batch size 128, AdamW optimizer, cosine learning rate schedule (2k-step warmup to $10^{-4}$, decay to $10^{-6}$), and weight decay $10^{-6}$. Training completes in approximately one day on an RTX 5090.

\begin{figure*}[t]
    \centering
    \includegraphics[width=0.09\linewidth]{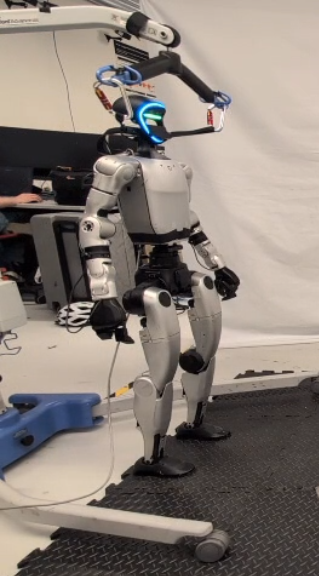}%
    \includegraphics[width=0.09\linewidth]{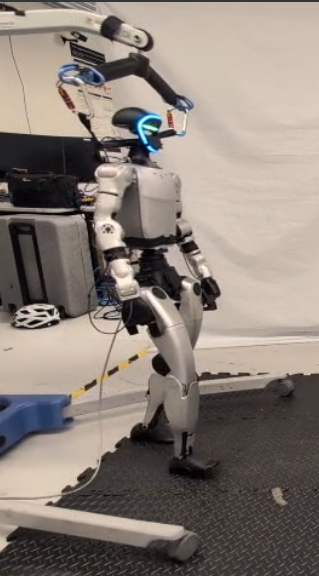}%
    \includegraphics[width=0.09\linewidth]{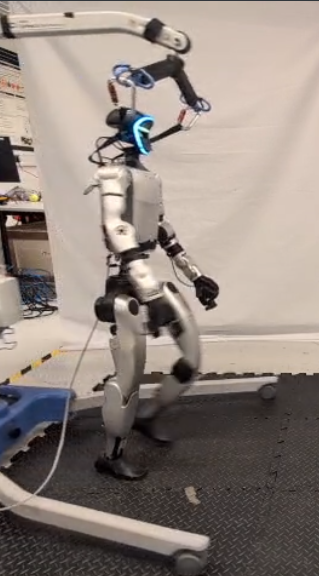}%
    \includegraphics[width=0.09\linewidth]{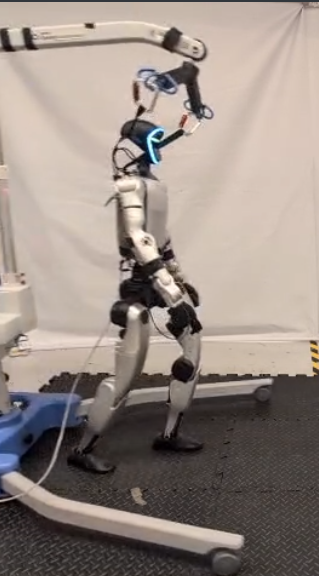}%
    \includegraphics[width=0.09\linewidth]{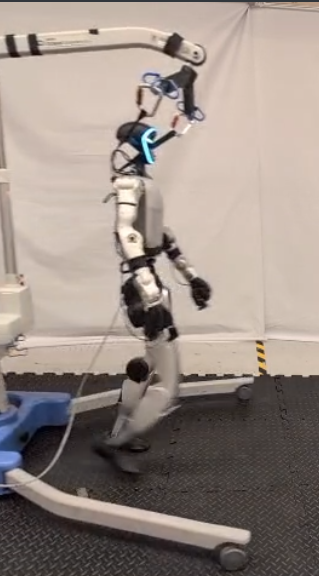}%
    \includegraphics[width=0.09\linewidth]{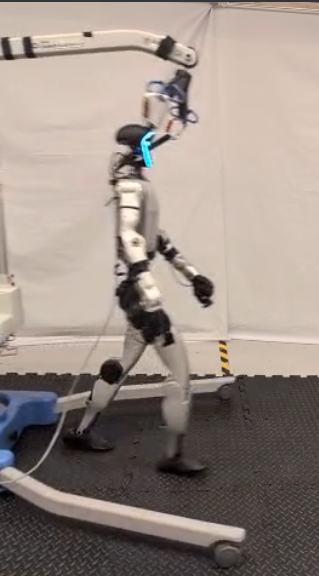}%
    \includegraphics[width=0.09\linewidth]{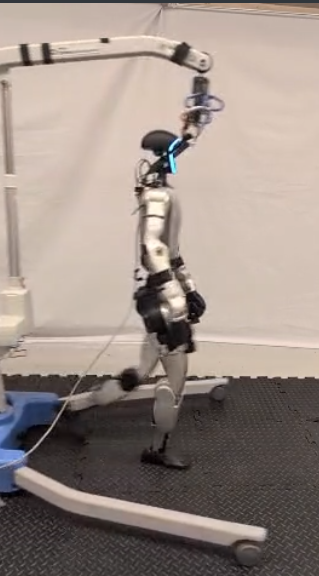}%
    \includegraphics[width=0.09\linewidth]{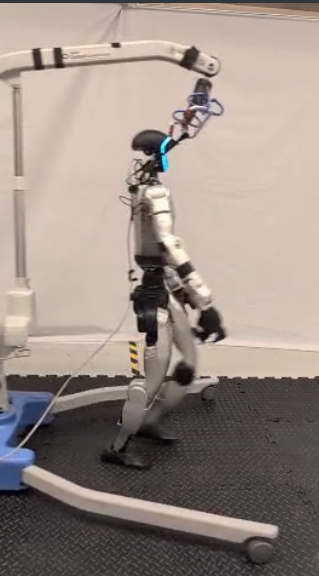}%
    \includegraphics[width=0.09\linewidth]{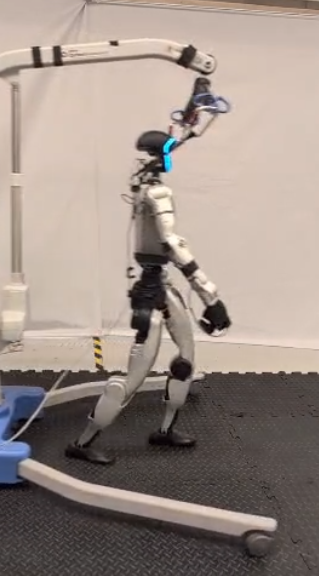}%
    \includegraphics[width=0.09\linewidth]{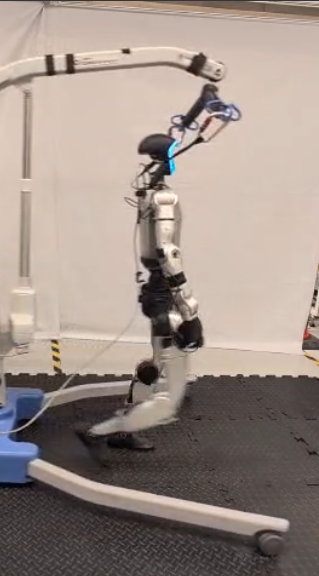}%
    \includegraphics[width=0.09\linewidth]{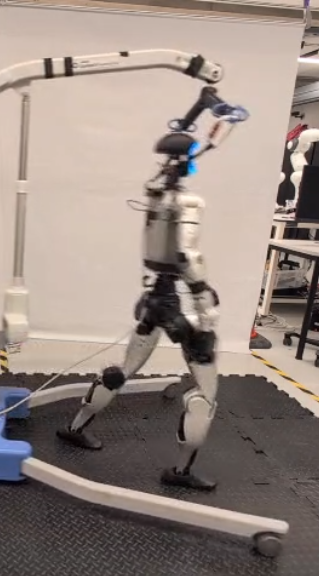}%

    \caption{
    \textbf{Fast forward walking behaviour.}
    Sequential frames showing the policy executing forward locomotion.
    }
    \label{fig:walk_fast_sequence}
\end{figure*}

\begin{figure*}[t]
    \centering

    \begin{minipage}[t]{0.49\linewidth}
        \centering
        \includegraphics[width=0.16\linewidth]{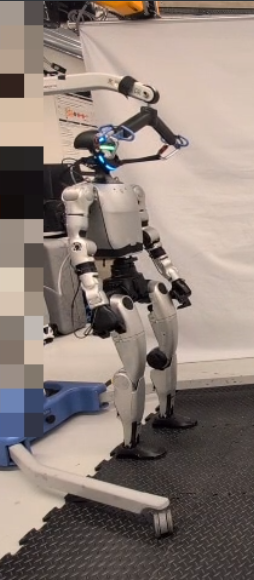}%
        \includegraphics[width=0.16\linewidth]{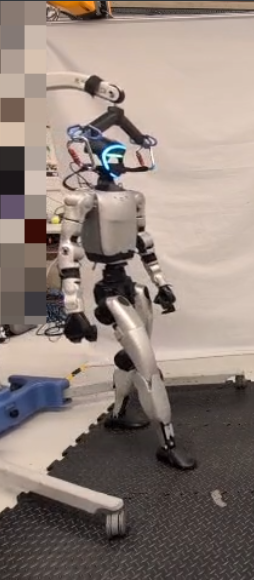}%
        \includegraphics[width=0.16\linewidth]{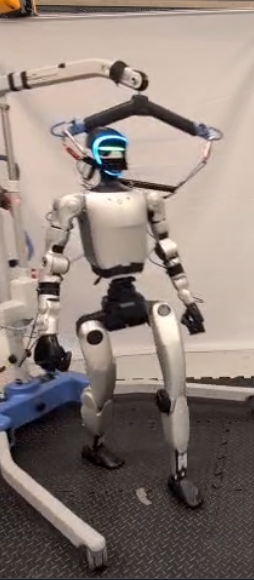}%
        \includegraphics[width=0.16\linewidth]{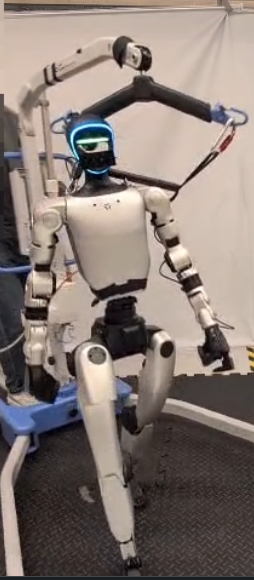}%
        \includegraphics[width=0.16\linewidth]{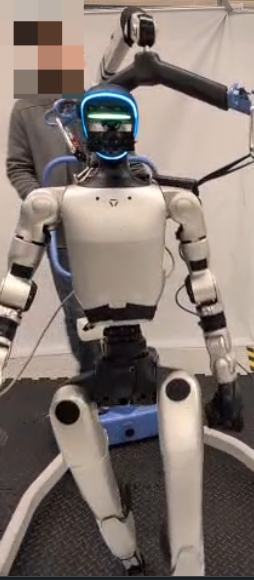}%
        \includegraphics[width=0.16\linewidth]{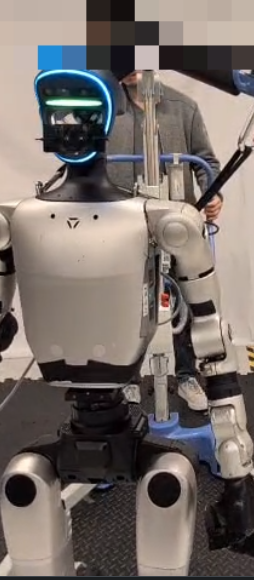}

        \caption*{(a) Time-lapsed frames showing \textbf{walking right}.}
    \end{minipage}
    \hfill
    \begin{minipage}[t]{0.49\linewidth}
        \centering
        \includegraphics[width=0.16\linewidth]{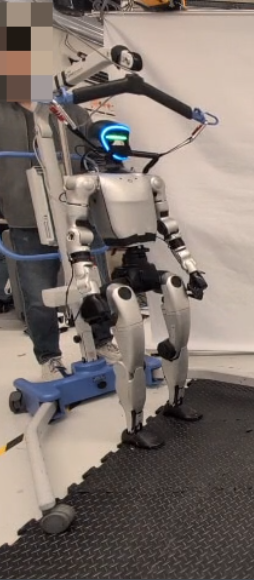}%
        \includegraphics[width=0.16\linewidth]{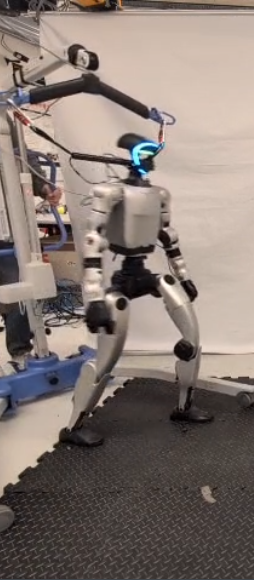}%
        \includegraphics[width=0.16\linewidth]{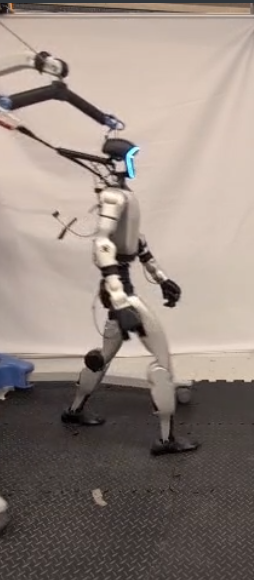}%
        \includegraphics[width=0.16\linewidth]{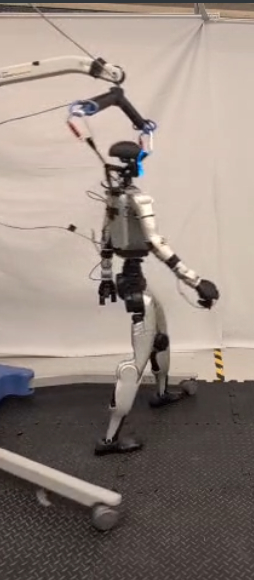}%
        \includegraphics[width=0.16\linewidth]{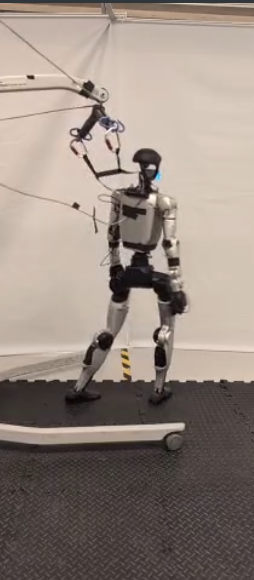}%
        \includegraphics[width=0.16\linewidth]{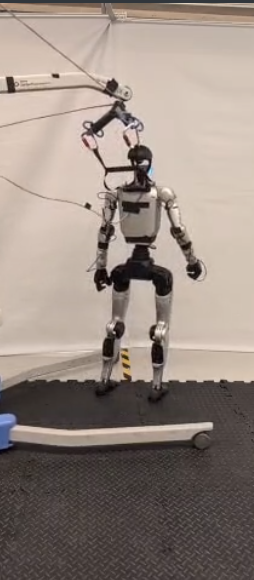}
        \caption*{(b) Time-lapsed frames showing \textbf{walking left}.}
    \end{minipage}

    \caption{Comparison of walking right and walking left sequences.}
    \label{fig:walking_sequences}
\end{figure*}

We deploy SCDP on a Unitree G1 at 50~Hz with the model running on a remote workstation (RTX 5090), achieving 105~Hz throughput via ONNX Runtime (Figs.~\ref{fig:walk_fast_sequence}, and \ref{fig:walking_sequences}).

\section{Simulation Results}
\label{sec:simulation_results}

\begin{figure*}[t]
    \centering
    \includegraphics[width=0.11\linewidth]{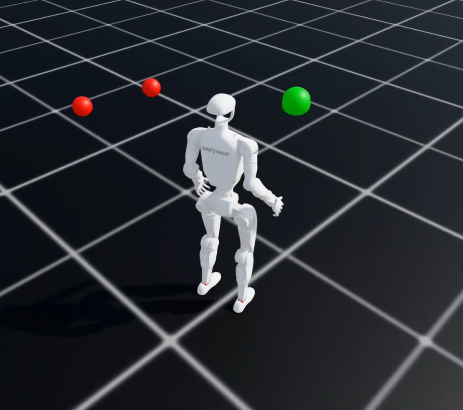}%
    \includegraphics[width=0.11\linewidth]{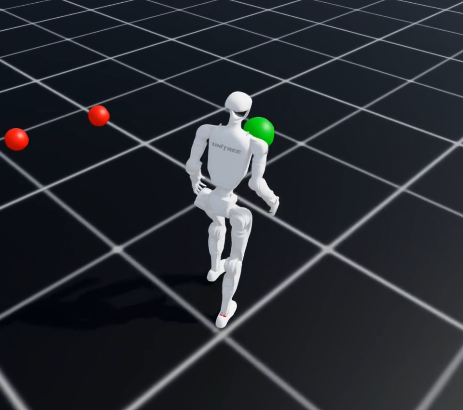}%
    \includegraphics[width=0.11\linewidth]{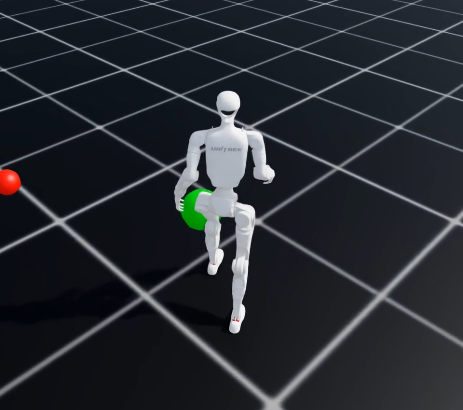}%
    \includegraphics[width=0.11\linewidth]{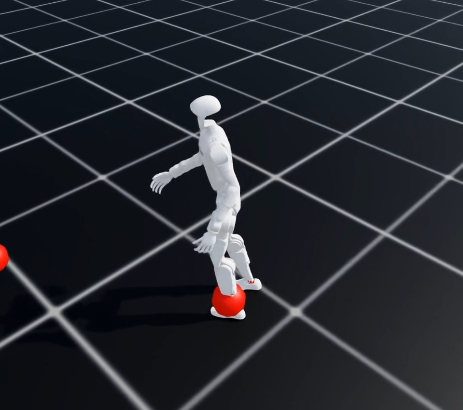}%
    \includegraphics[width=0.11\linewidth]{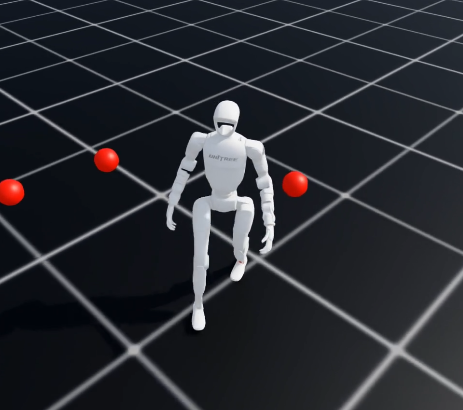}%
    \includegraphics[width=0.11\linewidth]{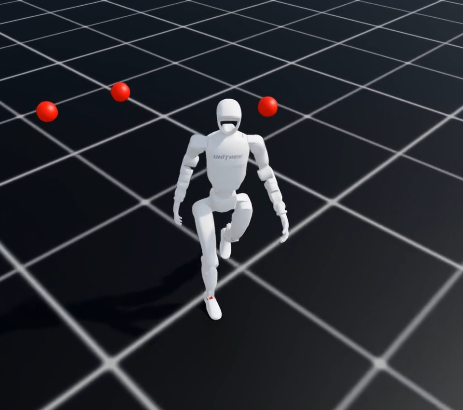}%
    \includegraphics[width=0.11\linewidth]{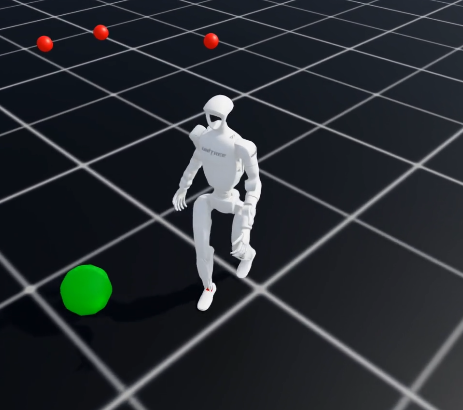}%
    \includegraphics[width=0.11\linewidth]{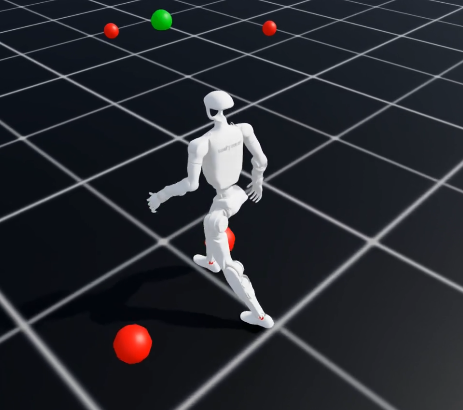}%
    \includegraphics[width=0.11\linewidth]{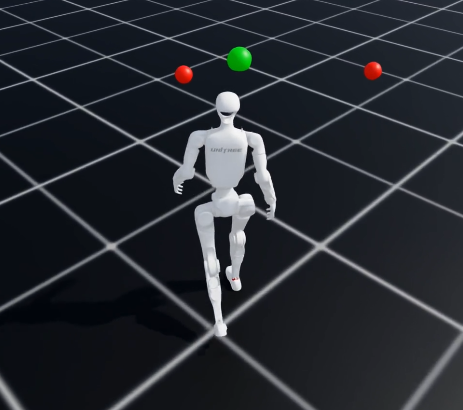}%

    \includegraphics[width=0.11\linewidth]{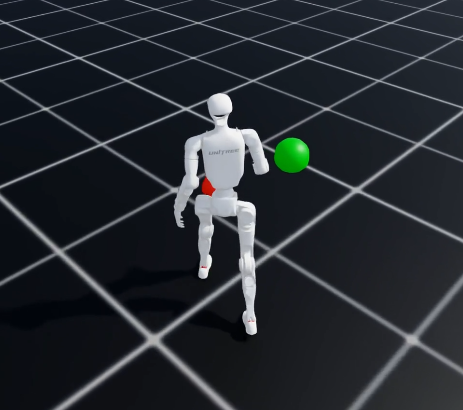}%
    \includegraphics[width=0.11\linewidth]{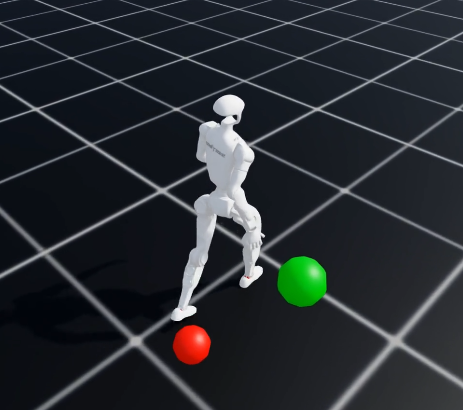}%
    \includegraphics[width=0.11\linewidth]{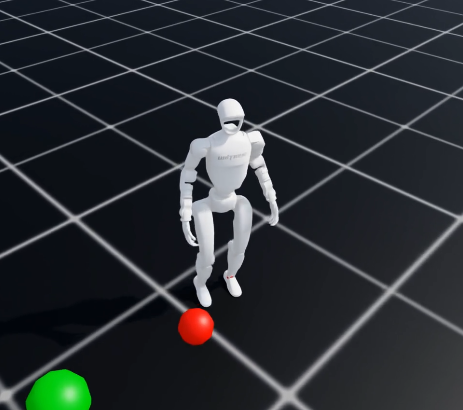}%
    \includegraphics[width=0.11\linewidth]{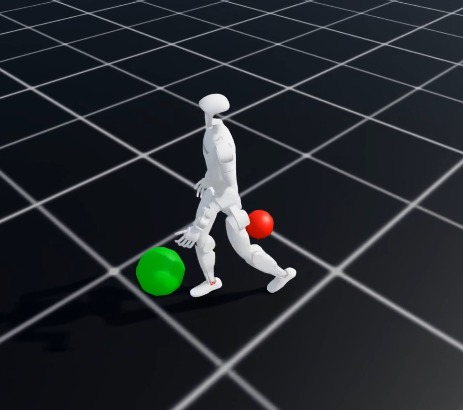}%
    \includegraphics[width=0.11\linewidth]{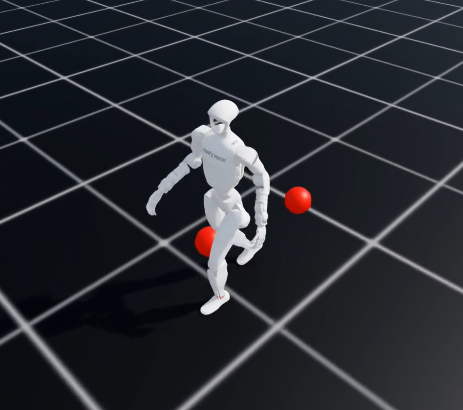}%
    \includegraphics[width=0.11\linewidth]{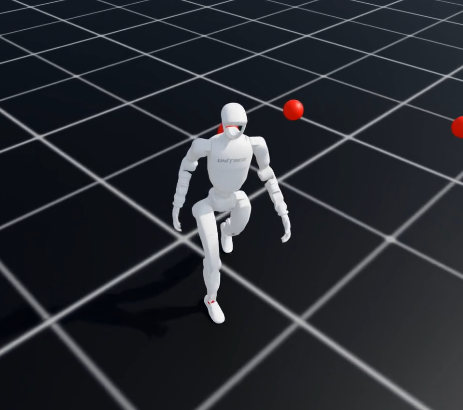}%
    \includegraphics[width=0.11\linewidth]{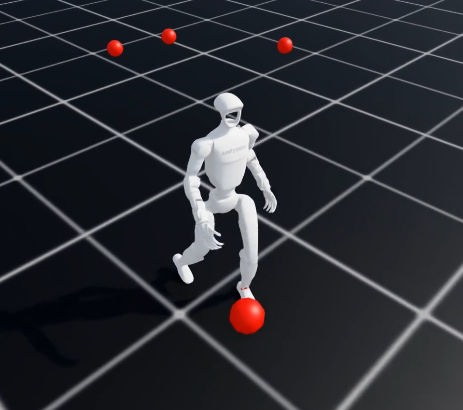}%
    \includegraphics[width=0.11\linewidth]{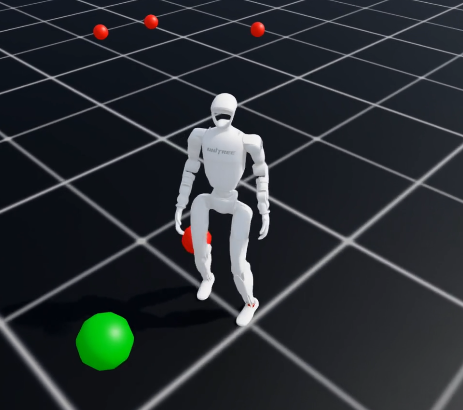}%
    \includegraphics[width=0.11\linewidth]{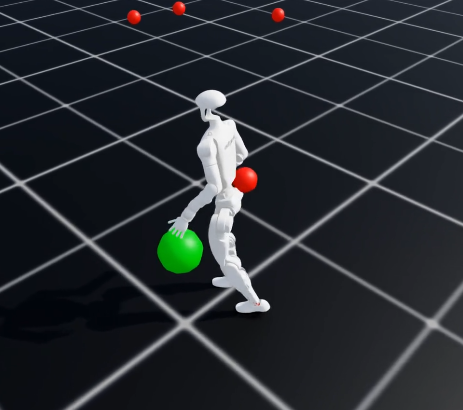}%
    
    \caption{
    \textbf{Waypoint navigation task.}
    Sequential frames showing the SCDP controlling the robot to navigate succesfully to all five waypoints (green: current target, red: remaining targets).
}
    \label{fig:navigation_sequence}
\end{figure*}

    
    


We evaluate the framework in simulation by addressing four questions on sensor-based diffusion control: (\textbf{Q1}) Can diffusion policies succeed with geometrically uninformative observations? (\textbf{Q2}) Can they achieve velocity control without state estimation? (\textbf{Q3}) What design choices are most critical? (\textbf{Q4}) Can SCDP transfer to motion tracking control?

All policies are evaluated at 50~Hz in IsaacLab~\cite{isaaclab} with domain randomization. We assess four tasks:
\textbf{(1) Perturbation Recovery:} The robot walks forward at 0.5\,m/s for 15 seconds while experiencing random velocity perturbations of $\pm 0.5$\,m/s every 1-3 seconds. Success requires maintaining balance throughout (1000 trials). Learning to recover from such perturbations is a fundamental motor skill that shall be acquired, as recent work has shown that such behaviors can be learned via RL \cite{fall_recovery_2023}.
\textbf{(2) Joystick Control:} The robot executes a sequence of discrete velocity commands (forward 0.5\,m/s, backward $-0.5$\,m/s, turn left $+1.0$\,rad/s, turn right $-1.0$\,rad/s), each held for 3 seconds. Success requires completing all commands without falling (100 trials).
\textbf{(3) Waypoint Navigation:} Five targets are randomly sampled within a radius of 2.5\,m. The robot must reach all waypoints within 60 seconds (Fig.~\ref{fig:navigation_sequence}). We report the mean completion rate (100 trials).
\textbf{(4) Motion Reference Tracking:} The robot reproduces specific reference motions. We evaluate using the Mean Per-Joint Position Error (MPJPE-L/G for local/global), velocity and acceleration distributions, and the success rate.

\begin{figure*}[t]
    \centering
    \begin{subfigure}{0.48\linewidth}
        \centering
        \includegraphics[width=\linewidth]{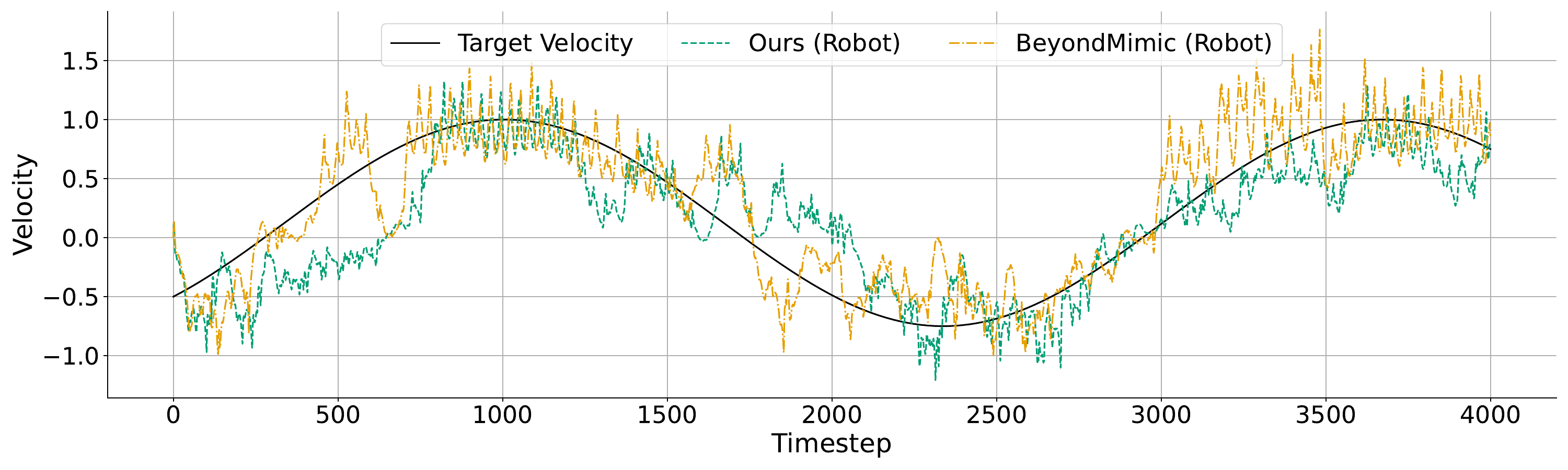}
        \subcaption{}
        \label{fig:vel_track_x}
    \end{subfigure}\hfill
    \begin{subfigure}{0.48\linewidth}
        \centering
        \includegraphics[width=\linewidth]{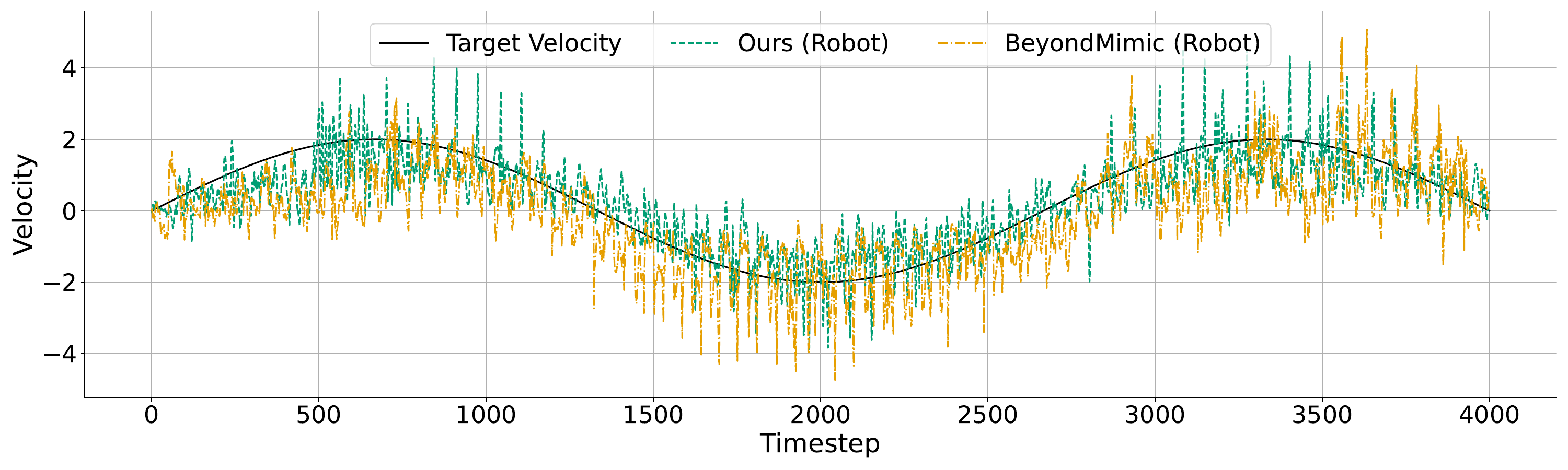}
        \subcaption{}
        \label{fig:vel_track_z}
    \end{subfigure}
    \caption{
    \textbf{Velocity tracking performance.} 
    Comparison of realized velocities against commanded targets (black) for (a) linear velocity $v_x$ and (b) yaw rate $\dot{\psi}$. SCDP (orange) exhibits smoother tracking with reduced oscillations compared to the BeyondMimic baseline (teal), with slight lag during direction changes.
}
    \label{fig:velocity_tracking}
\vspace{-4mm}
\end{figure*}

\subsection{Comparison with Baseline Methods}
\label{sec:main_results}

\begin{table}[t]
\centering
\scriptsize
\caption{Success/completion rates (\%) across observation modes. Subscript -- priv: privileged states; co: context observations only; mo: mixed observations. Bold: best per task.}
\label{tab:main_results}
\resizebox{\columnwidth}{!}{
\begin{tabular}{lcccc}
    \toprule
    \textbf{Method} & \textbf{$v_{\text{pelvis}}$ feedback} & \textbf{Perturb} & \textbf{Navigation} & \textbf{Joystick} \\
    \midrule
    BC$_{\text{co}}$     & $\checkmark$ & 70.1 & 43.8 & 70.0 \\
    BC$_{\text{co}}$     & $\times$ & 5.0 & 4.0 & 4.0 \\
    \midrule
    BeyondMimic$_{\text{priv}}$     & $\checkmark$ & 99.4 & 98.6 & \textbf{100.0} \\
    BeyondMimic$_{\text{co}}$       & $\checkmark$ & 20.7 & 22.2 & 3.0 \\
    BeyondMimic$_{\text{mo}}$      & $\checkmark$ & 99.4 & 94.4 & 94.0 \\
    BeyondMimic$_{\text{mo}}$      & $\times$ & 92.0 & 5.6 & 93.0 \\
    \midrule
    Ours$_{\text{priv}}$         & $\checkmark$ & 99.3 & \textbf{100.0} & \textbf{100.0} \\
    Ours$_{\text{co}}$           & $\checkmark$ & 59.8 & 58.8 & 11.0 \\
    Ours$_{\text{mo}}$          & $\checkmark$ & 99.1 & 99.4 & \textbf{100.0} \\
    Ours$_{\text{mo}}$     & $\times$ & 96.0 & 93.2 & 90.0 \\
    \midrule
    Ours: SCDP        & $\times$ & \textbf{99.8} & 99.8 & \textbf{100.0} \\
    \bottomrule
\end{tabular}}
\end{table}

We compare our approach against the classifier-guided baseline from~\cite{diffusecloc} and behavior cloning (BC). We evaluate four observation variants: \textbf{priv} ($s_t$ for context and prediction), \textbf{co} ($o_t$ for context and prediction), \textbf{mo} (mixed-observation training: $o_t$ for context, $s_t$ for prediction), and whether the context contains $v_{pelvis}$ feedback. 

Our final model, SCDP, employs mixed observation training, restricted denoising, context distribution alignment, and context-aware attention masking, operating without $v_{pelvis}$ feedback.

Table~\ref{tab:main_results} reveals several key findings. First, contextual-only policies (co) show significant degradation compared to their privileged counterparts, confirming that geometric information is essential in standard diffusion distillation (\textbf{Q1}). Mixed-observation training (mo) resolves this dependency entirely, with near-perfect performance comparable to their privileged counterparts. This demonstrates that diffusion policies can successfully operate from geometrically uninformative observations when optimized under our proposed mixed-observation training (\textbf{Q1}). 
Crucially, velocity feedback is essential for reliable control across prior methods. The results show that our full method, SCDP, does not depend on velocity feedback; it achieves competitive scores across all tasks, matching or exceeding the privileged methods. These results confirm that our proposed framework enables robust velocity control without the need for state estimation (\textbf{Q2}), enabling control under partial observability where previous methods fail.

\newcommand{\heat}[1]{%
  \begingroup
  \edef\val{#1}%
  \ifdim \val pt < 20pt \cellcolor{red!100}\else
  \ifdim \val pt < 30pt \cellcolor{red!85!green}\else
  \ifdim \val pt < 40pt \cellcolor{red!70!green}\else
  \ifdim \val pt < 50pt \cellcolor{red!60!green}\else
  \ifdim \val pt < 70pt \cellcolor{red!50!green}\else
  \ifdim \val pt < 75pt \cellcolor{red!40!green}\else
  \ifdim \val pt < 86pt \cellcolor{red!30!green}\else
  \ifdim \val pt < 90pt \cellcolor{red!20!green}\else
  \ifdim \val pt < 95pt \cellcolor{red!10!green}\else
  \ifdim \val pt < 99pt \cellcolor{green!90!red}\else
  \cellcolor{green!100}\fi\fi\fi\fi\fi\fi\fi\fi\fi\fi
  #1%
  \endgroup
}

\begin{table}[t]
\centering
\scriptsize
\setlength{\tabcolsep}{3pt}
\caption{Perturbation recovery success rates (\%) under varying data collection strategies,  Color intensity indicates performance (green=high, red=low).}
\label{tab:ds-ablation}
\begin{tabular}{c|cc|cc|cc}
\toprule
\textbf{Noise} 
  & \multicolumn{2}{c|}{\textbf{0.0 m/s}}
  & \multicolumn{2}{c|}{\textbf{0.5 m/s}} 
  & \multicolumn{2}{c}{\textbf{1.0 m/s}} \\
\textbf{Level}
  & \textbf{Push} & \textbf{No Push}
  & \textbf{Push} & \textbf{No Push}
  & \textbf{Push} & \textbf{No Push}\\
\midrule
0.0  & \heat{98.2} & \heat{32.5} & \heat{98.2} & \heat{24.7} & \heat{95.8} & \heat{18.7} \\
0.06 & \heat{99.3} & \heat{78.4} & \heat{99.0} & \heat{79.4} & \heat{98.9} & \heat{74.1} \\
0.12 & \heat{87.1} & \heat{81.6} & \heat{99.5} & \heat{98.0} & \heat{98.8} & \heat{95.0} \\
0.18 & \heat{97.5} & \heat{82.4} & \heat{99.4} & \heat{99.2} & \heat{98.8} & \heat{99.6} \\
0.24 & \heat{88.2} & \heat{70.0} & \heat{99.1} & \heat{99.6} & \heat{99.2} & \heat{99.3} \\
\bottomrule
\end{tabular}
\end{table}

\begin{table}[t]
\centering
\scriptsize
\setlength{\tabcolsep}{4pt}
\caption{Effect of context length on performance without velocity feedback. Bold indicates best per task.}
\label{tab:cl-ablation}
\begin{tabular}{lccc}
\toprule
\textbf{Context Length } & \textbf{$v_{\text{pelvis}}$ feedback} & \textbf{Perturb Succ. (\%)} & \textbf{Nav. Compl. (\%)}\\
\midrule
\multicolumn{4}{l}{\textbf{Ours$_{\text{mo}}$} } \\
\quad 4  & $\times$ & 84.3 & 74.2 \\
\quad 8  & $\times$ & 98.2  & 87.8 \\
\quad 16 & $\times$ & 55.5  & 33.4 \\
\midrule
\multicolumn{4}{l}{\textbf{SCDP}}\\
\quad 4  & $\times$ & \textbf{99.5} & \textbf{97.8}\\
\quad 8  & $\times$ & 96.8 & 86.6 \\
\quad 16 & $\times$ & 96.0  & 75.4 \\
\bottomrule
\end{tabular}
\vspace{-3mm}
\end{table}

\subsection{Data Collection Strategy Ablation}
\label{sec:data_ablation}
We vary the application of action noise (0.0--0.24 std) and push force during data collection, evaluating perturbation recovery at 0.0, 0.5, and 1.0 m/s. Table~\ref{tab:ds-ablation} shows that moderate action noise is essential. Without noise or pushing, policies achieve $<33\%$ success, improving to $>70\%$ with minimal noise. Second, push perturbations provide benefits without added noise, increasing the success from $<33\%$ to $>90\%$.
Applying both is optimal, exposing the policy to recovery behaviors that action noise alone cannot provide is beneficial, while excessive noise (0.24) degrades stationary performance (\textbf{Q3}).

\subsection{Context Length Ablation}
\label{sec:context_ablation}
We investigate how context length affects policy robustness without $v_{\text{pelvis}}$ feedback. Specifically, we evaluate context lengths of 4, 8, and 16, comparing SCDP with $\text{Ours}_{\text{mo}}$.
Table~\ref{tab:cl-ablation} shows that a four-step context sufficiently enables the SCDP model to capture dynamics, with longer histories degrading performance.
In contrast, Ours$_{\text{mo}}$ shows inconsistent performance across context lengths, with a non-monotonic relationship between context length and success rate. Notably, both methods degrade with 16-step contexts, suggesting that longer histories amplify compounding errors at inference or distribution mismatch (\textbf{Q3}).

\begin{table*}[!t]
\centering
\caption{Motion tracking performance on AMASS-walk dataset. MMP: expert upper bound.}
\label{tab:metrics_amass}
\resizebox{\textwidth}{!}{%
\begin{tabular}{lc|ccccc|ccccc}
\toprule
& & \multicolumn{5}{c|}{\textbf{Train}} & \multicolumn{5}{c}{\textbf{Test}} \\
\cmidrule(lr){3-7} \cmidrule(lr){8-12}
\textbf{Method} & \textbf{$v_{\text{pelvis}}$ feedback} & \textbf{MPJPE-G $\downarrow$} & \textbf{MPJPE-L $\downarrow$} & \textbf{Vel-Dist $\downarrow$} & \textbf{Accel-Dist $\downarrow$} & \textbf{Success $\uparrow$} & \textbf{MPJPE-G $\downarrow$} & \textbf{MPJPE-L $\downarrow$} & \textbf{Vel-Dist $\downarrow$} & \textbf{Accel-Dist $\downarrow$} & \textbf{Success $\uparrow$} \\
\midrule
MMP {\textit{(upper bound)}} & $\checkmark$ & 69.84 $\pm$ 25.96 & 47.73 $\pm$ 23.82 & 3.03 $\pm$ 1.07 & 1.00 $\pm$ 0.47 & 98.94 & 69.22 $\pm$ 26.12 & 46.58 $\pm$ 22.99 & 3.07 $\pm$ 1.04 & 1.03 $\pm$ 0.46 & 99.03 \\
\midrule
$\text{BC}$ & $\times$ & 370.36 $\pm$ 278.93 & 53.65 $\pm$ 23.15 & 12.87 $\pm$ 7.59 & 14.78 $\pm$ 13.91 & 30.19 & 371.66 $\pm$ 276.42 & \textbf{50.71} $\pm$ 19.66 & 6.39 $\pm$ 3.01 & \textbf{1.31} $\pm$ 0.55 & 31.24 \\
$\text{Ours}_{mo}$ & $\times$ & 457.87 $\pm$ 374.15 & 54.81 $\pm$ 22.82 & 6.16 $\pm$ 2.33 & \textbf{1.33} $\pm$ 0.48 & 78.39 & 473.91 $\pm$ 408.60 & 54.36 $\pm$ 21.98 & 6.41 $\pm$ 2.52 & 1.39 $\pm$ 0.51 & 79.17 \\
$\text{SCDP}$ & $\times$ & \textbf{282.43} $\pm$ 167.54 & \textbf{53.63} $\pm$ 23.60 & \textbf{5.37} $\pm$ 2.14 & 1.48 $\pm$ 0.70 & \textbf{93.12} & \textbf{288.30} $\pm$ 184.22 & 53.12 $\pm$ 23.03 & \textbf{5.55} $\pm$ 2.25 & 1.51 $\pm$ 0.74 & \textbf{93.22} \\
\bottomrule
\end{tabular}%
}
\vspace{-4mm}
\end{table*}

\begin{table}[t]
\centering
\scriptsize
\caption{Ablation study on key design components for diffusion policies without velocity feedback.}
\label{tab:feature-ablation}
\resizebox{\columnwidth}{!}{
\begin{tabular}{ccccccc}
\toprule
\textbf{Mixed Obs.} & \textbf{Representation} & \textbf{Context Dist.} & \textbf{Full Prefix} & \textbf{Restricted} & \textbf{Navigation} \\
\textbf{Training} & \textbf{Alignment} & \textbf{Alignment} & \textbf{Attention} & \textbf{Denoising} & \textbf{(\%)} \\
\midrule
\multicolumn{6}{l}{\textit{All features}} \\
$\checkmark$ & $\checkmark$ & $\checkmark$ & $\checkmark$ & $\checkmark$ & 89.2 \\
\multicolumn{6}{l}{\textit{No features }} \\
$\times$ & $\times$ & $\times$ & $\times$ & $\times$ & 1.4 \\
\midrule
\multicolumn{6}{l}{\textit{Section 1: Exclusion Ablation (all except one)}} \\
$\times$ & $\times$ & $\checkmark$ & $\checkmark$ & $\checkmark$ & 1.4 \\
$\checkmark$ & $\times$ & $\checkmark$ & $\checkmark$ & $\checkmark$ & \textbf{97.8} \\
$\checkmark$ & $\checkmark$ & $\times$ & $\checkmark$ & $\checkmark$ & 9.7 \\
$\checkmark$ & $\checkmark$ & $\checkmark$ & $\times$ & $\checkmark$ & 81.0 \\
$\checkmark$ & $\checkmark$ & $\checkmark$ & $\checkmark$ & $\times$ & 92.2 \\
\midrule
\multicolumn{6}{l}{\textit{Section 2: Single Feature (only one)}} \\
$\checkmark$ & $\times$ & $\times$ & $\times$ & $\times$ & 74.2 \\
$\checkmark$ & $\checkmark$ & $\times$ & $\times$ & $\times$ & 61.2 \\
$\times$ & $\times$ & $\checkmark$ & $\times$ & $\times$ & 1.6 \\
$\times$ & $\times$ & $\times$ & $\checkmark$ & $\times$ & 2.6 \\
$\times$ & $\times$ & $\times$ & $\times$ & $\checkmark$ & 1.2 \\
\bottomrule
\end{tabular}}
\vspace{-4mm}
\end{table}

\subsection{Feature Design Ablation}
\label{sec:feature_ablation}

We systematically evaluate the contribution of each design component when training without velocity feedback. We present results across four sets of configurations in Tab.~\ref{tab:feature-ablation}: all features enabled, no features, an exclusion ablation (all except one), and an inclusion ablation (only one feature). The exclusion ablation reveals that mixed-observation training is the foundation: removing it causes complete failure (1.4\%). Conversely, removing representation alignment achieves the best performance (97.8\%), showing that representation alignment interferes with optimization.
Context distribution alignment proves critical: its removal drops performance to 9.7\%, showing its importance for achieving sensor-based control without velocity feedback. The single-feature results demonstrate that mixed-observation training alone achieves 74.2\% success, while context distribution alignment, full prefix attention, and restricted denoising provide negligible benefits in isolation. These results show that mixed-observation training is necessary for sensor-based control, while context distribution alignment, full prefix attention, and restricted denoising collectively enable the full method to match privileged-observation performance, providing additional reliability and robustness (\textbf{Q3}). 
\vspace{-2mm}

\subsection{Velocity Tracking Performance}
\label{sec:velocity_tracking}
We evaluated velocity tracking fidelity and responsiveness compared to BeyondMimic~\cite{beyondmimic} and the classifier-guided baseline, while tracking commanded velocities varying between $\pm 1$\,m/s (linear) and $\pm 2$\,rad/s (angular). Both policies successfully track nominal velocities, but our method achieves tighter linear tracking with reduced oscillations (Fig.~\ref{fig:velocity_tracking}a), though with slightly increased lag ($\sim$250--500 ms) during direction reversals. Angular velocity tracking is comparable (Fig.~\ref{fig:velocity_tracking}b).
\vspace{-2mm}

\subsection{Motion Reference Tracking Performance}
\label{sec:motion_tracking}

We evaluate SCDP's ability to reproduce reference motions from its training distribution (AMASS).
Table~\ref{tab:metrics_amass} shows that each distillation method achieves comparable local pose accuracy (MPJPE-L) to the expert upper bound, measured while the episode has not terminated. Diffusion-based distillation training with mixed observation training already significantly outperforms the BC baseline, with an evaluation set success rate of 79\% compared to 31\%. Furthermore, SCDP greatly outperforms both methods with a success rate of 93\% and a reduction of the global position error (MPJPE-G) from 473 to 288, confirming its effectiveness at reference motion tracking (\textbf{Q4}). SCDP has significant global position error as opposed to the expert. This is expected because the expert observes its global drift from the reference and is able to correct it, while this is not possible under partial observability. SCDP must track the global position from the kinematic reference command alone.
\vspace{-2mm}




\section{Conclusion}
\label{sec:conclusion}
We presented SCDP that achieves diffusion distillation of whole-body humanoid controllers under partial observability via three techniques: mixed-observation training, restricted denoising, and context alignment.

Our study showed that previous diffusion and offline distillation methods fail with geometrically uninformative and partial observations, critically depending on the \textit{base velocity} feedback, while SCDP excels in this scenario. Furthermore, we highlight SCDP’s versatility, demonstrating the effectiveness in both velocity control (99\%) and motion reference tracking (93\%). 

We presented systematic ablations that identify the optimal design choices and data collection procedures. Finally, SCDP was running in a closed-loop at 50~Hz for performance validation on a real physical G1 robot, showing  \textit{robust sim-to-real transfer}. 

Future work shall explore sim-to-real fine-tuning that do not require privileged states, drift-correction mechanisms for long-horizon motion tracking, and the extension to contact-rich manipulation and complex terrain scenarios.

\addtolength{\textheight}{-14cm}   








\bibliographystyle{IEEEtran}
\bibliography{references}

@IEEEtranBSTCTL{IEEEexample:BSTcontrol,
  CTLuse_forced_etal       = "yes",
  CTLmax_names_forced_etal = "1",
  CTLnames_show_etal       = "1",
  CTLdash_repeated_names = "no",
}

@article{Bao_2025,
   title={Deep reinforcement learning for robotic bipedal locomotion: a brief survey},
   volume={59},
   ISSN={1573-7462},
   url={http://dx.doi.org/10.1007/s10462-025-11451-z},
   DOI={10.1007/s10462-025-11451-z},
   number={1},
   journal={Artificial Intelligence Review},
   publisher={Springer Science and Business Media LLC},
   author={Bao, Lingfan and Humphreys, Joseph and Peng, Tianhu and Zhou, Chengxu},
   year={2025},
   month=dec }

@misc{bao_2025_2,
      title={Hierarchical Intention-Aware Expressive Motion Generation for Humanoid Robots}, 
      author={Lingfan Bao and Yan Pan and Tianhu Peng and Dimitrios Kanoulas and Chengxu Zhou},
      year={2025},
      eprint={2506.01563},
      archivePrefix={arXiv},
      primaryClass={cs.RO},
      url={https://arxiv.org/abs/2506.01563}, 
}

@inproceedings{
mdm,
title={Human Motion Diffusion Model},
author={Guy Tevet and Sigal Raab and Brian Gordon and Yoni Shafir and Daniel Cohen-or and Amit Haim Bermano},
booktitle={The Eleventh International Conference on Learning Representations },
year={2023},
url={https://openreview.net/forum?id=SJ1kSyO2jwu}
}

@INPROCEEDINGS{Tianhu2,
  author={Peng, Tianhu and Bao, Lingfan and Zhou, Chengxu},
  booktitle={2025 IEEE-RAS 24th International Conference on Humanoid Robots (Humanoids)}, 
  title={Gait-Conditioned Reinforcement Learning with Multi-Phase Curriculum for Humanoid Locomotion}, 
  year={2025},
  volume={},
  number={},
  pages={1-7},
  doi={10.1109/Humanoids65713.2025.11203058}}

@misc{dagger,
      title={A Reduction of Imitation Learning and Structured Prediction to No-Regret Online Learning}, 
      author={Stephane Ross and Geoffrey J. Gordon and J. Andrew Bagnell},
      year={2011},
      eprint={1011.0686},
      archivePrefix={arXiv},
      primaryClass={cs.LG},
      url={https://arxiv.org/abs/1011.0686}, 
}

@article{diffpolicy,
  title={Diffusion policy: Visuomotor policy learning via action diffusion},
  author={Chi, Cheng and Xu, Zhenjia and Feng, Siyuan and Cousineau, Eric and Du, Yilun and Burchfiel, Benjamin and Tedrake, Russ and Song, Shuran},
  journal={The International Journal of Robotics Research},
  pages={02783649241273668},
  year={2023},
  publisher={SAGE Publications Sage UK: London, England}
}

@article{diffusecloc,
  title={Diffuse-CLoC: Guided Diffusion for Physics-based Character Look-ahead Control},
  author={Huang, Xiaoyu and Truong, Takara and Zhang, Yunbo and Yu, Fangzhou and Sleiman, Jean Pierre and Hodgins, Jessica and Sreenath, Koushil and Farshidian, Farbod},
  journal={ACM Transactions on Graphics (TOG)},
  volume={44},
  number={4},
  pages={1--13},
  year={2025},
  publisher={ACM},
  note={SIGGRAPH 2025}
}

@InProceedings{diffuceloco,
  title = 	 {DiffuseLoco: Real-Time Legged Locomotion Control with Diffusion from Offline Datasets},
  author =       {Huang, Xiaoyu and Chi, Yufeng and Wang, Ruofeng and Li, Zhongyu and Peng, Xue Bin and Shao, Sophia and Nikolic, Borivoje and Sreenath, Koushil},
  booktitle = 	 {Proceedings of The 8th Conference on Robot Learning},
  pages = 	 {1567--1589},
  year = 	 {2025},
  editor = 	 {Agrawal, Pulkit and Kroemer, Oliver and Burgard, Wolfram},
  volume = 	 {270},
  series = 	 {Proceedings of Machine Learning Research},
  month = 	 {06--09 Nov},
  publisher =    {PMLR},
  url = 	 {https://proceedings.mlr.press/v270/huang25a.html}
}

@misc{birodiff,
      title={BiRoDiff: Diffusion policies for bipedal robot locomotion on unseen terrains}, 
      author={GVS Mothish and Manan Tayal and Shishir Kolathaya},
      year={2024},
      eprint={2407.05424},
      archivePrefix={arXiv},
      primaryClass={cs.RO},
      url={https://arxiv.org/abs/2407.05424}, 
}

@inproceedings{pdp,
author = {Truong, Takara Everest and Piseno, Michael and Xie, Zhaoming and Liu, Karen},
title = {PDP: Physics-Based Character Animation via Diffusion Policy},
year = {2024},
isbn = {9798400711312},
publisher = {Association for Computing Machinery},
address = {New York, NY, USA},
url = {https://doi.org/10.1145/3680528.3687683},
doi = {10.1145/3680528.3687683},
booktitle = {SIGGRAPH Asia 2024 Conference Papers},
articleno = {86},
numpages = {10},
location = {Tokyo, Japan},
series = {SA '24}
}

@article{margolis2022walktheseways,
    title={Walk These Ways: Tuning Robot Control for Generalization with Multiplicity of Behavior},
    author={Margolis, Gabriel B and Agrawal, Pulkit},
    journal={Conference on Robot Learning},
    year={2022}
}

@InProceedings{learntowalk,
  title = 	 {Learning to Walk in Minutes Using Massively Parallel Deep Reinforcement Learning},
  author =       {Rudin, Nikita and Hoeller, David and Reist, Philipp and Hutter, Marco},
  booktitle = 	 {Proceedings of the 5th Conference on Robot Learning},
  pages = 	 {91--100},
  year = 	 {2022},
  editor = 	 {Faust, Aleksandra and Hsu, David and Neumann, Gerhard},
  volume = 	 {164},
  series = 	 {Proceedings of Machine Learning Research},
  month = 	 {08--11 Nov},
  publisher =    {PMLR},
  url = 	 {https://proceedings.mlr.press/v164/rudin22a.html}
}

@article{deepmimic,
author = {Peng, Xue Bin and Abbeel, Pieter and Levine, Sergey and van de Panne, Michiel},
title = {DeepMimic: example-guided deep reinforcement learning of physics-based character skills},
year = {2018},
issue_date = {August 2018},
publisher = {Association for Computing Machinery},
address = {New York, NY, USA},
volume = {37},
number = {4},
issn = {0730-0301},
url = {https://doi.org/10.1145/3197517.3201311},
doi = {10.1145/3197517.3201311},
journal = {ACM Trans. Graph.},
month = jul,
articleno = {143},
numpages = {14}
}

@inproceedings{onmih20,
  title={OmniH2O: Universal and Dexterous Human-to-Humanoid Whole-Body Teleoperation and Learning},
  author={He, Tairan and Luo, Zhengyi and He, Xialin and Xiao, Wenli and Zhang, Chong and Zhang, Weinan and Kitani, Kris and Liu, Changliu and Shi, Guanya},
  journal={arXiv preprint arXiv:2406.08858},
  year={2024}
}

@article{xbody2,
  title={ExBody2: Advanced Expressive Humanoid Whole-Body Control}, 
  author={Ji, Mazeyu and Peng, Xuanbin and Liu, Fangchen and Li, Jialong and Yang, Ge and Cheng, Xuxin and Wang, Xiaolong},
  journal={arXiv preprint arXiv:2412.13196},
  year={2024},
  }

@inproceedings{humanplus,
  author    = {Fu, Zipeng and Zhao, Qingqing and Wu, Qi and Wetzstein, Gordon and Finn, Chelsea},
  title     = {HumanPlus: Humanoid Shadowing and Imitation from Humans},
  booktitle = {{Conference on Robot Learning (CoRL)}},
  year      = {2024},
}

@article{twist,
title={TWIST: Teleoperated Whole-Body Imitation System},
author= {Yanjie Ze and Zixuan Chen and João Pedro Araújo and Zi-ang Cao and Xue Bin Peng and Jiajun Wu and C. Karen Liu},
year= {2025},
journal= {arXiv preprint arXiv:2505.02833}
}

@misc{unitracker,
      title={UniTracker: Learning Universal Whole-Body Motion Tracker for Humanoid Robots}, 
      author={Kangning Yin and Weishuai Zeng and Ke Fan and Minyue Dai and Zirui Wang and Qiang Zhang and Zheng Tian and Jingbo Wang and Jiangmiao Pang and Weinan Zhang},
      year={2025},
      eprint={2507.07356},
      archivePrefix={arXiv},
      primaryClass={cs.RO},
      url={https://arxiv.org/abs/2507.07356}, 
}

@article{gmt,
title={GMT: General Motion Tracking for Humanoid Whole-Body Control},
author={Chen, Zixuan and Ji, Mazeyu and Cheng, Xuxin and Peng, Xuanbin and Peng, Xue Bin and Wang, Xiaolong},
journal={arXiv:2506.14770},
year={2025}
}

@misc{beyondmimic,
      title={BeyondMimic: From Motion Tracking to Versatile Humanoid Control via Guided Diffusion}, 
      author={Qiayuan Liao and Takara E. Truong and Xiaoyu Huang and Guy Tevet and Koushil Sreenath and C. Karen Liu},
      year={2025},
      eprint={2508.08241},
      archivePrefix={arXiv},
      primaryClass={cs.RO},
      url={https://arxiv.org/abs/2508.08241}, 
}

@misc{guidedMDM,
      title={Guided Motion Diffusion for Controllable Human Motion Synthesis}, 
      author={Korrawe Karunratanakul and Konpat Preechakul and Supasorn Suwajanakorn and Siyu Tang},
      year={2023},
      eprint={2305.12577},
      archivePrefix={arXiv},
      primaryClass={cs.CV},
      url={https://arxiv.org/abs/2305.12577}, 
}

@inproceedings{
  closd,
  title={{CL}o{SD}: Closing the Loop between Simulation and Diffusion for multi-task character control},
  author={Guy Tevet and Sigal Raab and Setareh Cohan and Daniele Reda and Zhengyi Luo and Xue Bin Peng and Amit Haim Bermano and Michiel van de Panne},
  booktitle={The Thirteenth International Conference on Learning Representations},
  year={2025},
  url={https://openreview.net/forum?id=pZISppZSTv}
}

@article{maskedmimic,
    author = {
        Tessler, Chen and Guo, Yunrong and Nabati, Ofir and Chechik, Gal
        and Peng, Xue Bin
    },
    title = {
        MaskedMimic: Unified Physics-Based Character Control Through
        Masked Motion Inpainting
    },
    year = {2024},
    journal={ACM Transactions on Graphics (TOG)},
    publisher={ACM New York, NY, USA}
}

@software{mink,
  author = {Zakka, Kevin},
  title = {{Mink: Python inverse kinematics based on MuJoCo}},
  year = {2025},
  month = may,
  version = {0.0.11},
  url = {https://github.com/kevinzakka/mink},
  license = {Apache-2.0}
}

@article{ppo,
  title={Proximal policy optimization algorithms},
  author={Schulman, John and Wolski, Filip and Dhariwal, Prafulla and Radford, Alec and Klimov, Oleg},
  journal={arXiv preprint arXiv:1707.06347},
  year={2017}
}

@misc{ddpm,
      title={Denoising Diffusion Probabilistic Models}, 
      author={Jonathan Ho and Ajay Jain and Pieter Abbeel},
      year={2020},
      eprint={2006.11239},
      archivePrefix={arXiv},
      primaryClass={cs.LG},
      url={https://arxiv.org/abs/2006.11239}, 
}

@conference{amass,
  title = {{AMASS}: Archive of Motion Capture as Surface Shapes},
  author = {Mahmood, Naureen and Ghorbani, Nima and Troje, Nikolaus F. and Pons-Moll, Gerard and Black, Michael J.},
  booktitle = {International Conference on Computer Vision},
  pages = {5442--5451},
  month = oct,
  year = {2019},
  month_numeric = {10}
}

@article{isaaclab,
   author={Mittal, Mayank and Yu, Calvin and Yu, Qinxi and Liu, Jingzhou and Rudin, Nikita and Hoeller, David and Yuan, Jia Lin and Singh, Ritvik and Guo, Yunrong and Mazhar, Hammad and Mandlekar, Ajay and Babich, Buck and State, Gavriel and Hutter, Marco and Garg, Animesh},
   journal={IEEE Robotics and Automation Letters},
   title={Orbit: A Unified Simulation Framework for Interactive Robot Learning Environments},
   year={2023},
   volume={8},
   number={6},
   pages={3740-3747},
   doi={10.1109/LRA.2023.3270034}
}

@InProceedings{classGuidance,
  title = 	 {Deep Unsupervised Learning using Nonequilibrium Thermodynamics},
  author = 	 {Sohl-Dickstein, Jascha and Weiss, Eric and Maheswaranathan, Niru and Ganguli, Surya},
  booktitle = 	 {Proceedings of the 32nd International Conference on Machine Learning},
  pages = 	 {2256--2265},
  year = 	 {2015},
  editor = 	 {Bach, Francis and Blei, David},
  volume = 	 {37},
  series = 	 {Proceedings of Machine Learning Research},
  address = 	 {Lille, France},
  month = 	 {07--09 Jul},
  publisher =    {PMLR},
  url = 	 {https://proceedings.mlr.press/v37/sohl-dickstein15.html}
}

@article{yang2020multi,
  title={Multi-expert learning of adaptive legged locomotion},
  author={Yang, Chuanyu and Yuan, Kai and Zhu, Qiuguo and Yu, Wanming and Li, Zhibin},
  journal={Science Robotics},
  volume={5},
  number={49},
  pages={eabb2174},
  year={2020},
  publisher={American Association for the Advancement of Science}
}

@inproceedings{distilling_2024,
  title={Distilling reinforcement learning policies for interpretable robot locomotion: Gradient boosting machines and symbolic regression},
  author={Acero, Fernando and Li, Zhibin},
  booktitle={2024 IEEE/RSJ International Conference on Intelligent Robots and Systems (IROS)},
  pages={6840--6847},
  year={2024},
  organization={IEEE}
}

@article{fall_recovery_2023,
  title={Learning complex motor skills for legged robot fall recovery},
  author={Yang, Chuanyu and Pu, Can and Xin, Guiyang and Zhang, Jie and Li, Zhibin},
  journal={IEEE Robotics and Automation Letters},
  volume={8},
  number={7},
  pages={4307--4314},
  year={2023},
  publisher={IEEE}
}

\end{document}